\documentclass[letterpaper, 10 pt, conference]{ieeeconf}  % Comment this line out if you need a4paper

\IEEEoverridecommandlockouts                              % This command is only needed if 
\usepackage{graphicx}
\usepackage{adjustbox}
\usepackage{multirow}
\usepackage{amssymb}
\usepackage{color}
\usepackage{tikz}
\usepackage{pgfplotstable}
\usepackage{pgfplots}
\usepackage[normalem]{ulem}

\usepackage{cite}

\usepackage{lineno}
\usepackage{subfig}
\usepackage{tabularx, rotating, caption}
\usepackage{array}
\usepackage{siunitx}
\usepackage{url}
\usepackage{booktabs}
\usepackage{comment}
\usepackage{cleveref}
\usepackage{pifont}% http://ctan.org/pkg/pifont
\newcommand{\rose}{ROSE}
\newcommand{\voronoi}{\texttt{Voronoi}}
\newcommand{\morph}{\texttt{Morph}}
\newcommand{\dist}{\texttt{Dist}}
\newcommand{\ours}{ROSE$^2$}

\title{\Large \bf 
Robust Structure Identification and Room Segmentation\\ of Cluttered Indoor Environments from Occupancy Grid Maps}

\author{Matteo Luperto$^{*1}$, Tomasz Piotr Kucner$^{2,3}$, Andrea Tassi$^{4}$, Martin Magnusson$^{2}$,\\  and Francesco Amigoni$^{4}$% <-this % stops a space
\thanks{1: Università degli studi di Milano, Italy.}
\thanks{2: Örebro Universitet, Sweden.} 
\thanks{3: Aalto University, Finland.} 
\thanks{4: Politecnico di Milano, Italy.}
\thanks{$^{*}$ Corresponding Author: {\tt\small matteo.luperto@unimi.it}
}
}

\begin{document}

\maketitle
\thispagestyle{empty}
\pagestyle{empty}

%%%%%%%%%%%%%%%%%%%%%%%%%%%%%%%%%%%%%%%%%%%%%%%%%%%%%%%%%%%%%%%%%%%%%%%%%%%%%%%%
\begin{abstract}
Identifying the environment's structure, i.e., to detect core components as rooms and walls, can facilitate several tasks fundamental for the successful operation of indoor autonomous mobile robots, including semantic environment understanding. %\rem{, localisation, task and motion planning.}
%\rem{The existing algorithms rely on the fact that one of the key feature of human-made indoor environments is a high level of structure.
%It is caused by a prevalent use of straight lines in connection with right angles in the design of it.}
These robots often rely on 2D occupancy maps for core tasks such as localisation, motion and task planning.
%\addt{Occupancy maps are widely used because they are easily obtained, and resilient to different events common in dynamic settings.}
%
%However, reliable identification of structure from 2D occupancy maps is still an open problem due to clutter (e.g., furniture and movable objects) that is perceived by the robots' sensors and occludes the structure (e.g., walls). 
However, reliable identification of structure and room segmentation from 2D occupancy maps is still an open problem due to clutter (e.g., furniture and movable object), occlusions, and partial coverage. 
We propose a method for the RObust StructurE identification and ROom SEgmentation (\ours ) of 2D occupancy maps, which may be cluttered and incomplete.
\ours{} identifies the main directions of walls and is resilient to clutter and partial observations, allowing to extract a clean, abstract geometrical  floor-plan-like description of the environment, which is used to segment, i.e., to identify rooms in, the original occupancy grid map.
\ours{} is tested in several real-world publicly-available cluttered maps obtained in different conditions. The results show how it can robustly identify the environment structure in 2D occupancy maps suffering from clutter and partial observations, while significantly  improving room segmentation accuracy. Thanks to the combination of clutter removal and robust room segmentation \ours{} consistently achieves higher performance than the state-of-the-art methods, against which it is compared.
\end{abstract}

\section{Introduction}\label{sec:INTRO}

\begin{figure}
 \centering
    	\subfloat[Occupancy grid map.\label{fig:1A}]{ \includegraphics[width=0.68\linewidth]{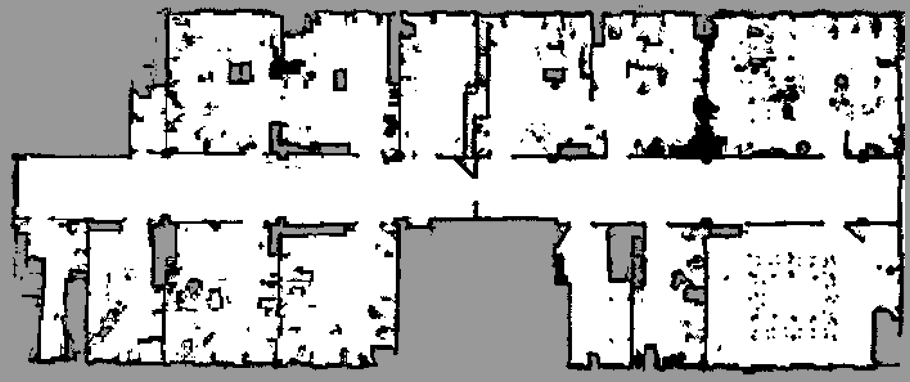}}\\	
	\subfloat[Removal of non-structural components.\label{fig:1B}]{ \includegraphics[width=0.68\linewidth]{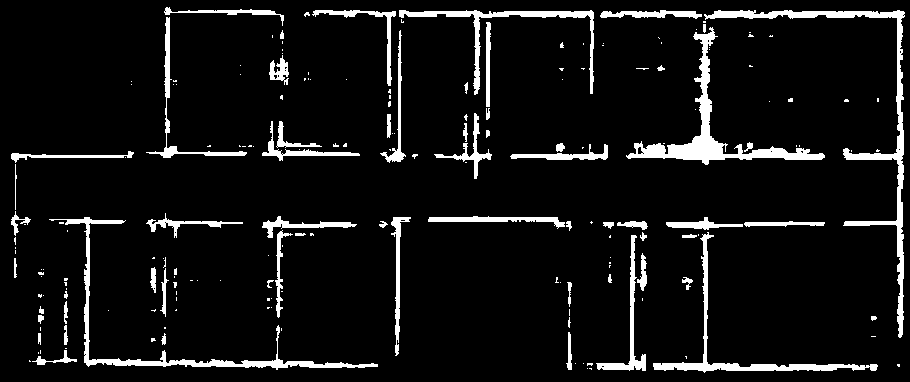}}\\
    	\subfloat[Wall lines identification.\label{fig:1C}]{ \includegraphics[width=0.68\linewidth]{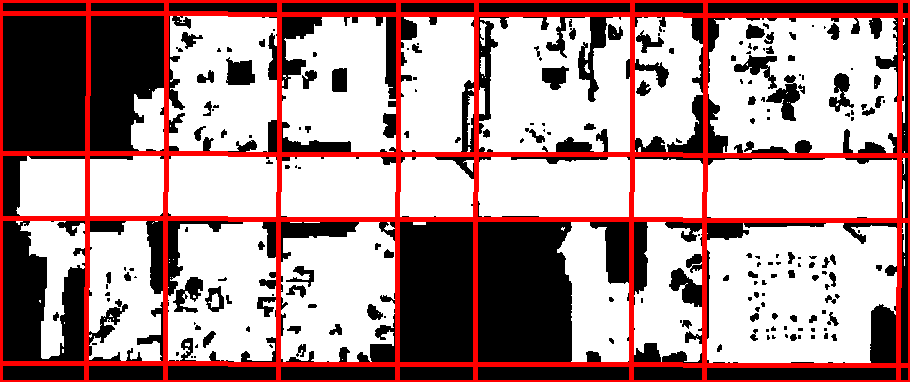}}\\	
    	\subfloat[Geometric floor-plan-like representation.\label{fig:1D}]{ \includegraphics[width=0.68\linewidth]{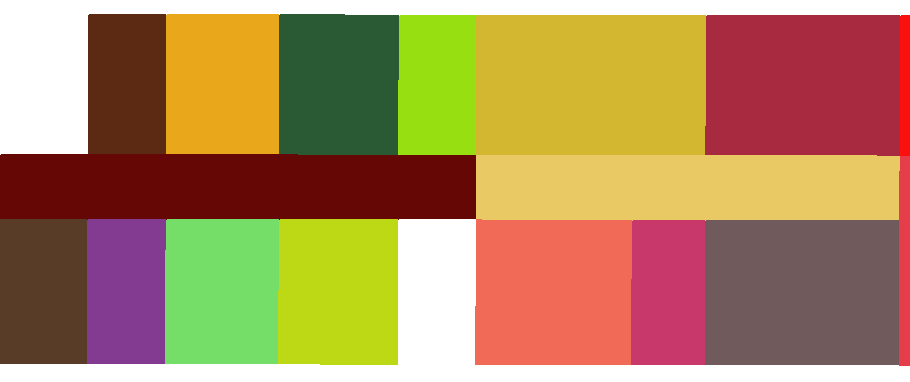}}\\
	 \subfloat[Room segmentation and room-shape prediction.\label{fig:1E}]{ \includegraphics[width=0.68\linewidth]{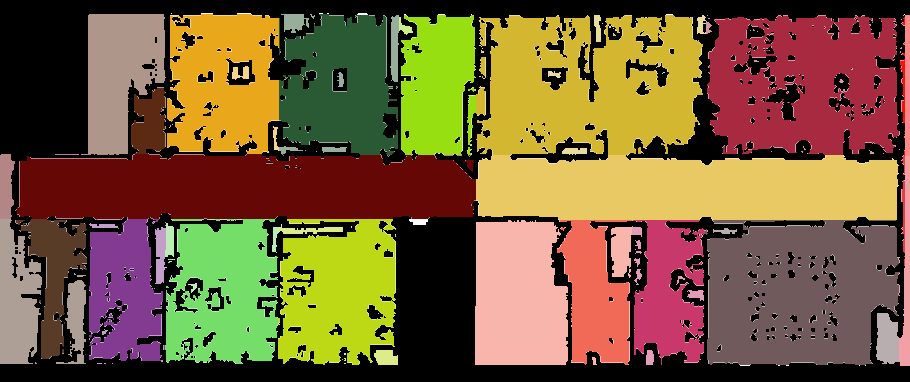}}
    	\\	
 \caption{Our method (\ours{}) starts from a (cluttered or partial) 2D occupancy grid map of an indoor environment and identifies the structure by retrieving its rooms and walls. To do so, it filters clutter and non-structural features (b), it retrieves the main wall directions (c), and obtains a geometrical representation of the environment (d), that is used to segment the map in rooms while predicting their shape (e).\label{fig:EXE}} 
\end{figure}

In recent years, ground mobile robots have been deployed in numerous indoor applications including industrial, public, office, and domestic environments. These robots often rely on 2D occupancy maps for core robotic tasks such as localization, motion and task planning.

Occupancy maps represent the shape of the environment through a grid, where each cell is associated to a probability of being occupied by an obstacle \cite{moravec1985}.
However, such maps do not provide higher-level semantic information about the environment (i.e., the type of objects, the structure of the environment, the room type~\cite{martinez2007supervised,magnusson-2017-places}).

Considering how semantic information about structural layout plays an important role in numerous robotic applications such as localisation \cite{structurelocalization22,howard2021lalaloc}, exploration \cite{AAMAS21,RAS2020,li2015semantically}, mapping performance benchmarking \cite{IROS18},  and room segmentation \cite{bormann2016room}, it is not surprising that the structure extraction problem has received substantial attention recently.
Even though indoor environments are predominantly well \emph{structured} it is still challenging to extract such information from 2D occupancy maps.
This is usually caused by the presence of clutter (e.g., from furniture, movable objects) that occludes the background (e.g., walls) \cite{ambrucs2014meta}.
Several methods that try to identify structure from 2D occupancy maps (e.g., by segmenting the map into rooms) perform poorly in cluttered environments \cite{mielle2018method}, as they are based on local map features such as corners and narrow passages. However, 2D occupancy maps provide little insights to distinguish between structural components of the environment (e.g., a wall) or furniture (e.g., a chair or a table), that can both give rise to such local features.
Some works tackled this problem by enriching the 2D range data with additional modalities like camera images or by using 3D point clouds  \cite{ambrucs2014meta, ARMENI}. 
However, additional sensing modalities are not always available.

In this work, we propose a method for RObust StructurE identification and ROom SEgmentation (\ours) from (cluttered) 2D occupancy maps. 
This method stems from the assumption that the structure of the environment can be defined through a (limited) set of main directions that are followed by walls.

The proposed method for structure extraction and room segmentation consists of two principal steps, as shown in Fig. \ref{fig:EXE}. First, we identify all occupied cells of a 2D occupancy map that belong to the main structure (e.g., walls), by using a robust frequency-based structure extraction method \cite{ICRA21}. At the same time, we filter out map components that are due to clutter, noise, and non-structural components.

Second, we group locally-perceived portions of walls along the identified main directions of walls in the environment, following the assumption that a wall could be shared by multiple rooms. 
This allows us to align walls detected from individual rooms at the building level.
We use these features to build a clean, abstract geometrical floor-plan-like-representation of the environment. 

The structure identified by our method may be used for multiple tasks such as  (i) room segmentation, (ii) floor plan reconstruction, (iii) prediction of the missing portion of the environment, (iv) regularization, (v) map inpainting. In this work, we apply structural knowledge to improve performance of room segmentation, while we show qualitative results in tasks (ii)-(v). 

\ours{} is applicable to different types of maps and environments as it does not rely on any assumption about the shape of the environment (as (pseudo-)Manhattan worlds), map completion (complete or partial maps), nor assume any parametrization specific for a given map. \ours{} does not require any training before being applied to new settings and can be used online during robot activities. 

The experimental evaluation is performed by showing how our method effectively identifies the structure and performs room segmentation in several environments, both using real-world cluttered and partial occupancy maps and on the benchmark dataset for room segmentation of~\cite{bormann2016room}.
Thanks to the combination of clutter removal and robust room segmentation \ours{} consistently achieves a higher performance when compared with the three methods from the state-of-the-art discussed in \cite{bormann2016room}.ß

This paper builds on previously published work about different yet related topics, on structure extraction \cite{IAS15, ICRA21} and shape prediction of unobserved rooms \cite{ICRA2019}.
In this paper, we integrate an improved version of our work of \cite{ICRA2019} with the robust feature extraction method of \cite{ICRA21} to achieve a framework that could be robustly used on all types of 2D occupancy maps for structure extraction and room segmentation. This improves the overall performance of the framework in two ways. First, it simplifies the steps of the method and removes the requirement of a manual tuning of a set of parameters for each type of maps (as partial or complete maps, simulated or real ones). Second, we remove the limitation of \cite{ICRA2019} to work only with clean simulated maps while also significantly improving performance in detecting the environment's structure. 

\section{Related Work}
\label{sec:ART}

A key feature of indoor environments is that they are built with an underlying \emph{structure} that is due to the fact that they are composed of doors, walls, rooms, and hallways \cite{luperto2013}.
Identifying the environment' structure is a key factor for addressing tobot applications, among others, as localisation \cite{structurelocalization22,howard2021lalaloc} and exploration \cite{AAMAS21,RAS2020,li2015semantically}.
In this section, we provide an overview of methods proposed to detect structure from occupancy robot maps. In particular, we focus on the task of \emph{room segmentation}, i.e., the detection and segmentation of a map into a set of rooms.

The survey of Bormann et al.~\cite{bormann2016room} compares the state-of-the-art techniques used to perform room segmentation on 2D maps. In particular, \cite{bormann2016room} identifies three main groups of approaches. The first and most popular one is that of Voronoi-based approaches \cite{thrun1998learning}, that segment the environment using a Voronoi graph, a spatial partition of the map whose nodes and edges have a maximal distance from at least two points of a finite set of obstacles. The second group is that of morphological operators, which segment the environment by using a transform on the map. An example is the morphological segmentation method \cite{buschka2002virtual}, that grows iteratively obstacles in the map until two connected areas become separated. A second example is the distance transform-based segmentation \cite{diosi}, that uses the distance of each empty pixel to the closest border/obstacle for segmentation.
The third group, that includes \cite{martinez2007supervised}, relies on learning grid-cell labels from local appearances (e.g., by classifying 2D raw sensorial inputs) and harmonizing neighbouring labels afterward. In this way, the process of room segmentation is fused with that of semantic mapping (e.g., determining that a room is also a corridor or an office). The performance of all these methods, as we show in Section~\ref{sec:EXP}, is unstable when applied to cluttered maps.

Mielle et al. \cite{mielle2018method} presents a method for room segmentation of 2D maps using the layout of the free space, by detecting ripple-like patterns and by merging neighboring regions with similar values. However, such a method cannot be applied to cluttered and non-empty environments.

The work of \cite{foroughi2021mapsegnet} addresses the task of room segmentation of a 2D occupancy map as a computer vision problem using an encoder-decoder architecture to identify rooms and corridors. Differently from us, this method requires training on the type of maps to be segmented. Clutter maps, which can present significantly different features than those considered for training in \cite{foroughi2021mapsegnet}, were not investigated.

The method of \cite{liu2014generalizable}, starting from a 2D occupancy map, reconstructs the geometrical shape of rooms by using Markov Logic Networks and data-driven Markov Chain Monte Carlo (MCMC) to sample over several possible room shape candidates, and selecting the fittest candidate according to the sensor data and a probabilistic room model. 
%In their work, two type of maps are used: abstract maps of empty environments and detailed maps, i.e. annotated floor plan of environments representing also furniture. Clutter maps, which can present significantly different features were not investigated.}

Several methods extract the environment structure from 3D point clouds~\cite{ARMENI,ochmann2016automatic,ambrucs2017automatic,oesau2014indoor}. While being partly inspired by those methods, our work focuses on a different type of input as 2D occupancy maps.

The task of using knowledge obtained from camera images and 3D point clouds to perform structure detection, room segmentation, and semantic mapping on 2D maps has also been investigated in several works. The work of \cite{pronobis2012large} shows how the integration of heterogeneous multi-modal information from 2D laser range scanners and vision can be fused into a probabilistic framework to obtain a semantic segmentation of the map into rooms. %, where the presence of rooms and their label can be observed; the result is a a probabilistic graphical representation of the environment that could be used also to make inference about unseen features of the environment (e.g., unobserved rooms).} 

A similar approach is presented in~\cite{sunderhauf2016place} where a deep learning vision-based place categorization method provides a segmentation and semantic map of the 2D occupancy maps. Another recent work following a similar method is the one of \cite{SmeanticFusionICCAS}, which integrates a vision-based scene classifier and an object detector to segment 2D grid maps into rooms and, at the same time, distinguish the semantic labels of that environment.

The work of \cite{he2021hierarchical} addresses a different, but related, problem. Starting from a 3D point-cloud map of the environment the method extracts a 2D occupancy map where different rooms are detected. To do so, they identify in 3D several structural features of the environment as regions, volumes, and passages, that are used to extract a topological map. 

Our works of \cite{IAS15,ICRA2019} present a method to predict the shape of partially-observed rooms in 2D occupancy maps after computing a geometrical floor-plan-like representation of the environment called the building layout. While showing promising results in clean maps without furniture, the method required an ad hoc parametrization for each map in order to work, which jeopardizes its applicability to cluttered maps. Nevertheless, we show in \cite{AAMAS21} how such predicted structural knowledge could be used online to improve performance in exploration for map building. Another application is in \cite{ECMR21}, that shows how the identification of the structure of the environment can help to predict the shape of multiple closed rooms that are behind closed doors, then performing inpainting on the map of the predicted shape of unobserved closed rooms. % More precisely, the proposed approach  can provide a reliable estimation of the full map of the environment by relying on structural knowledge even in settings where a few rooms have been actually observed by the robot, and when up to 15 rooms are not observed (and thus predicted) as located behind a closed door.}

\section{Our Method}\label{sec:OUR}
\ours{} identifies the structure present in  indoor environments from %their 
2D occupancy maps to reconstruct the geometry of structural elements like walls and rooms, that are utilised for segmentation.
Our method is divided into three steps: map cleaning and structural features identification (\Cref{sec:FEAT}),  wall detection (\Cref{sec:WALLS}), and room detection (\Cref{sec:ROOMS}).

The proposed pipeline is designed to work with regular categorical 2D occupancy grid maps.
In this type of map each cell can belong to one of three categories: occupied, free, or unknown.
Besides this, our approach is independent of the configuration of the robot system and of the SLAM algorithm used to build the occupancy map.

\subsection{Map cleaning and structural features identification}
\label{sec:FEAT}

Typically, for ground robots, occupancy maps $M$ obtained in real-world environments are particularly noisy, since the lidar used as the main sensorial input to build them is commonly placed close to the floor level. 
As a result, the sensor perceives and adds to the map non-structural objects such as legs of tables and chairs, bags, noise due to reflections caused by mirrors, French windows, and glass walls. 
Consequently, detecting the structural features, like walls, in such maps could be challenging.

Thus, in the first step of the proposed pipeline we implement a method for differentiating between structural elements of the environment and noise (i.e., clutter, spurious measurements, and non-structural elements of the environment)
and further removing noise in order to obtain a \emph{clean map} $\bar{M}$. To do so, we rely on the method that we presented in \cite{ICRA21}, called \rose.

\rose{} exploits the fact that in human-made environments structural components, like walls, are organised along a limited number of \emph{dominant directions} $\Psi$. 

As a consequence the dominant directions can be observed in the frequency spectrum as sets of radial ridges passing through the center of the frequency image.
To identify these ridges \rose{} computes  a 2D Discrete Fourier Transform (DFT) of $M$. 
Then, lines corresponding to the dominant directions are identified.
The process starts by computing the cumulative amplitude along each direction of $\psi$  in the frequency image, then those directions that correspond to the most prominent peaks among all directions $\psi$ are selected as $\Psi$. 
These peaks define the components of the frequency spectrum that should be retained (structure), while the reminder of the frequency spectrum will be set to zero (clutter).
This process is equivalent to automatically generating a band pass filter.
However, in contrast to typical image processing% approach% 
, the goal is to retain high-energy parts of the spectrum.
Finally, through computation of the inverse DFT for the filtered frequency image,  each occupied cell in the map gets a \emph{score} denoting how much it contributes to the structure of the environment.

Note that \rose{} does not assume Manhattan environments (with only right angles), but can also handle environments with more than two dominant directions.
For full details about the method, please refer to \cite{ICRA21}.

Through discarding cells with a score lower than a threshold (thus identifying them as not belonging to the structural elements) the clean map $\bar{M}$ is generated. The threshold is automatically tuned by checking different threshold values for each map to have a ratio of line segments (obtained using the method from \cite{kiryati1991probabilistic}) per free grid map cells in $\bar{M}$ inside a desired interval (experimentally identified), to avoid any further parameters tuning when assessing a new map type.

\subsection{Wall detection}\label{sec:WALLS}

\begin{figure}
   \centering
  \subfloat[\label{fig:wallA}]{ \includegraphics[width=0.495\linewidth]{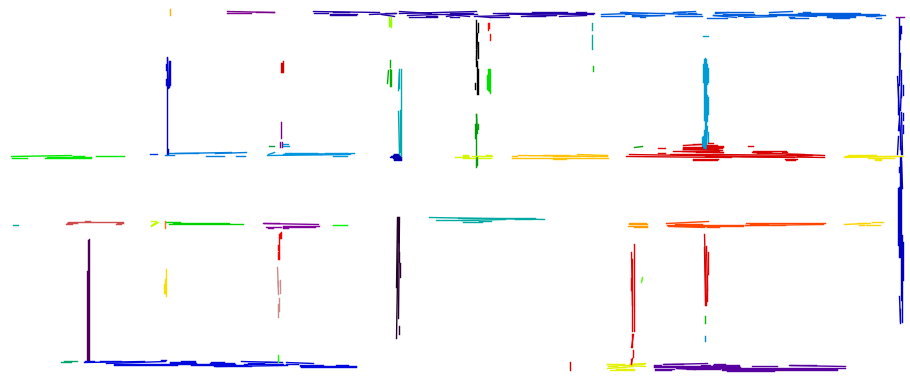}}
\subfloat[\label{fig:wallB}]{ \includegraphics[width=0.495\linewidth]{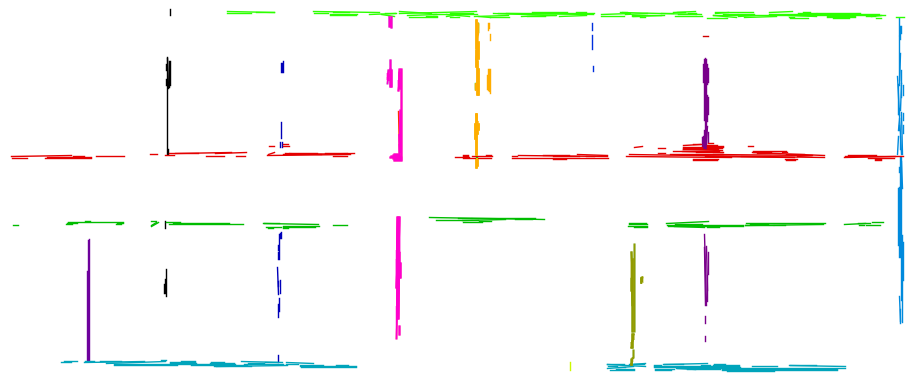}}
  \caption{ 
Line segments obtained from the clean map $\bar{M}$ of the map shown in Fig. \ref{fig:EXE}. Different clusters of collinear line segments are shown with different colors. Note how a single wall in the environment is represented as a series of slightly misaligned small line segments. Collinear clusters of (a) are merged in (b).}
  \label{fig:SPATIAL}
\end{figure}

While the process of obtaining a clear map $\bar{M}$ and  dominant directions $\Psi$ helps in identifying the main structural components, this is not sufficient to identify walls. 
This is because, in $\bar{M}$, portions of the same walls are usually represented by small line segments slightly misaligned w.r.t. each other.  
An example can be seen in Fig. \ref{fig:SPATIAL}, that shows the line segments obtained from  the map $\bar{M}$ of Fig. \ref{fig:1B}. It can be seen how, despite the clean map $\bar{M}$ (Fig. \ref{fig:1B}) just contains some clutter, line segments identified in it are (slightly) misaligned, despite belonging to the same wall.
%This knowledge is used to identify \emph{walls} within the map. 
To mitigate this limitation, we identify \emph{walls} by combining together line segments that are close to each other within the map. 
These line segments are clustered together according to their direction and to the fact that they are lying (or not) on one of the main dominant directions $\Psi$ identified in $\bar{M}$.

The first operation on $\bar{M}$ is the detection of line segments $s \in S$ using the probabilistic Hough line transform \cite{kiryati1991probabilistic}.
The line segments $S$ are then clustered together to identify those that belong to a portion of the same wall $W$. 
To do so, we rely on the method we described in \cite{ICRA2019}. 
At first, we cluster together all line segment with a similar angular coefficient.
After that, we identify a wall $W =\{s_1,\ldots,s_n\}$ as a cluster of line segments (obtained using DBSCAN, \cite{ester1996density}) that are spatially close to each other (i.e., there is a continuity between them) and have a similar angular coefficient. 
For full details about this step, please refer to \cite{ICRA2019}. 

We then align all line segment clusters $W$ to dominant directions $\Psi$, by associating each line segment $s\in W$ with its closest dominant direction $\psi$. 
Note that, as line segment $s$ are obtained from the clean map $\bar{M}$ from which $\psi$ is computed, all line segments follow one of the dominant directions. 
Thus, for each line segment $s$, we compute a line segment $\bar{s}$ which is a projection of $s$ on a line with direction $\psi$ passing through the  middle point $p$ of $s$, thus aligning all line segments in $\bar{M}$ to a dominant direction in $\Psi$.
As a result, for each wall cluster $W$ (of unaligned line segments $s$), we obtain the corresponding aligned  cluster $\bar{W}$ (of line segments $\bar{s}$ aligned with a dominant direction).

As walls are shared by different rooms, we have often the case where the same wall in the environment is split into multiple clusters $W$, as in the case of the walls along the main corridor of Fig. \ref{fig:wallA}. To solve this, we merge together collinear wall clusters $\bar{W}$ , using the following procedure.
Given a wall $\bar{W}=\{\bar{s}_1, \ldots,\bar{s}_n\}$, we represent it with its central point $P$. The central point $P$ is selected as the median point across all middle points $p_i$ of the line segments $\bar{s}_{i}\in \bar{W}$, along a direction perpendicular to the dominant direction of $W$.
Given two walls $\bar{W}$ and $\bar{W}'$, we call $l$ and $l'$ the parallel lines passing through the middle points $P$ and $P'$, both with dominant direction coefficient $\psi$. If the distance between $l$ and $l'$ is less than a threshold (intuitively, closer than the width of a doorway), then the two corresponding walls $\bar{W}$ and $\bar{W}'$ are merged together in $\bar{W}''$ (for which a new middle point $P''$ is computed). The result at the end of this process for Fig \ref{fig:wallA} is shown in Fig. \ref{fig:wallB}, where all line segments belonging to the same wall along the central corridor are assigned to the same cluster.

At this point, for each cluster $W_k$ we obtained, a \emph{representative line} $l_{k}$ is assigned as a line whose angular coefficient is equal to a dominant direction $\psi$ and passing through $p_k$. 
Each representative line, in red in Fig.~\ref{fig:1C}, indicates the direction of a wall within the building, and is projected across the whole map $M$. 

The convex area delimited by intersections of representative lines is called a \emph{face}.
We call $L$ the set of representative lines $l$, $F$ the set of faces $f$, and we define the edges $e\in E$ as the portions of representative lines common between two faces $f,f'$.
Note how there is a relation among all of these objects: a wall cluster $W$ is composed of line segments $s$; for each wall cluster $W$ exists a representative line $l$; $l$ indicates the direction of that wall $W$.

We define an edge weight $w(e)$ that represents how much of $e$ is ``covered'' by a projection on it of line segments $s$ of a wall $W$ corresponding to the representative line $l$ on which $e$ lies. Intuitively, if an edge $e$ is fully covered by  line segments (in $\bar{M})$, it has $w(e) \approx 1 $; an edge $e'$ located in an empty part of $\bar{M}$ has $w(e') \approx 0$. We perform then a filtering process: we remove all representative lines whose cumulative weight of their edges is below a threshold (empirically set to $0.1$, so that less than the $10\%$ of them is covered by an obstacle in $\bar{M}$) so to remove all representative lines that are caused by a local feature that may be not common to the whole environment. 
The rationale behind this is to identify a set of \emph{few meaningful} representative lines that cover the walls as observed by the robot in the map $M$. An example is shown in Fig. \ref{fig:1C}. 
To avoid to loose interesting local structural features with the filtering process, we keep only the edges $\underline{e}$ of a removed representative line $\underline{l}$ that are almost entirely covered (so with a large weight $w(\underline{e})$).

\subsection{Room detection}\label{sec:ROOMS}
The third and last step is to reconstruct the shape of the rooms in the environment.
Rooms are found by clustering faces according to the following two rules: (i) adjacent faces whose common edge corresponds to a wall should belong to different rooms, (ii) adjacent faces that share an edge not corresponding to any wall should be part of the same room. 
To do so, we partition $F$ in $\{F_1, F_2,\ldots\, F_{n}\}$, in which each $F_{i}$ collects the faces of a room, by clustering together faces using DBSCAN \cite{ester1996density} as the clustering method, and using the weight $w(e_{f,f'})$ (as defined in the previous section) of the edge adjacent to faces $f$ and $f'$ as the metric, similarly to what we did in \cite{ICRA2019}.  
 
The faces belonging to each $F_{i}$ are then merged together, obtaining a polygonal representation of the room $r_{i}$, in which the borders of the polygon are assumed to be the external walls of the room. 
The floor plan $\mathcal{F} =\{r_1,r_2, \ldots, r_{n} \}$ is finally retrieved by considering together all rooms $r$, as displayed in Fig. \ref{fig:1E}. 

The reconstruction of the floor plan $\mathcal{F}$ is a key step in our method. It extracts the aligned shape and location of the walls as observed by the robot and, simultaneously, the border of a room in $\mathcal{F}$ predicts the presence of walls that complete the rooms even when they are not observed (yet) by the robot (e.g., due to occlusion).  Note that the estimated floor plan can be used to infer the actual shape of a partially observed room, as in Fig. \ref{fig:1E}. This step differs from the one of \cite{ICRA2019}, where two different methods are used to first extract the shape of fully-mapped rooms \cite{IAS15}, and then to predict the shape of partially observed ones.

Finally, we obtain the segmented map $\check{M}$, where empty cells in the map $M$ are assigned to the corresponding room as identified in $\mathcal{F}$. 

The output of our method is thus the set $\langle \bar{M}, L, F, \mathcal{F}, \check{M} \rangle$ which identifies the environment's structure.

\subsection{Integrating missing structural knowledge}
\begin{figure}
 \centering
     \subfloat[\label{fig:explB}]{ \includegraphics[width=0.32\linewidth]{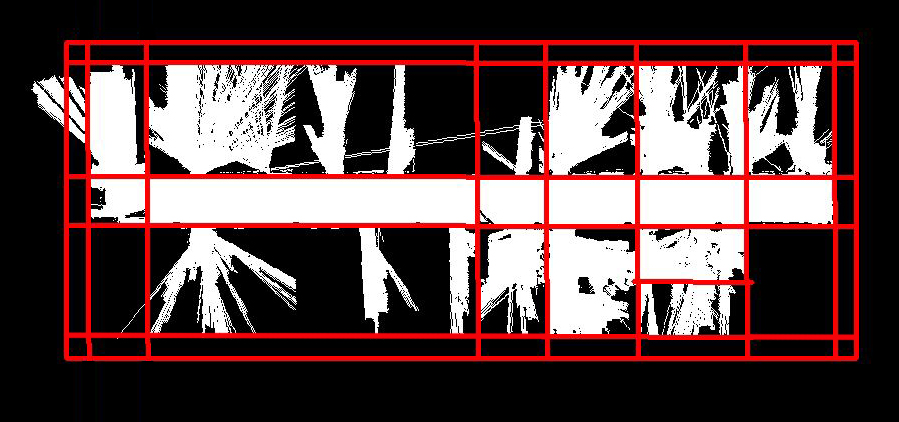}}
    \subfloat[\label{fig:VORONOI}]{  \includegraphics[width=0.32\linewidth]{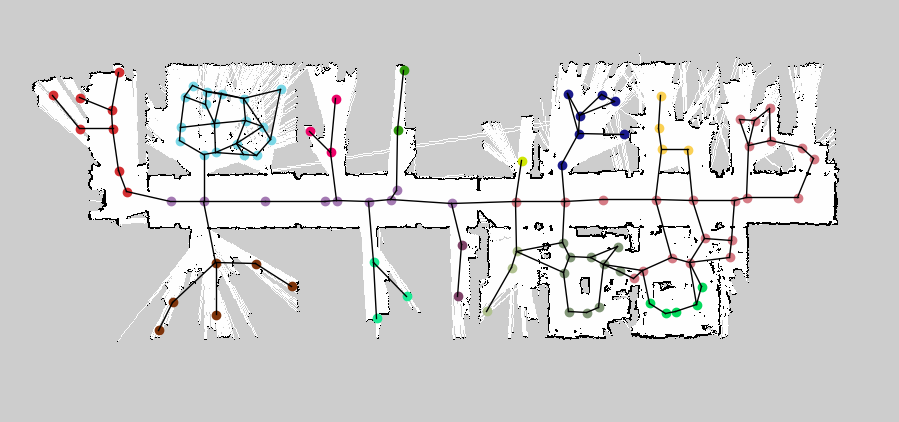}}
	\subfloat[\label{fig:explA}]{ \includegraphics[width=0.32\linewidth]{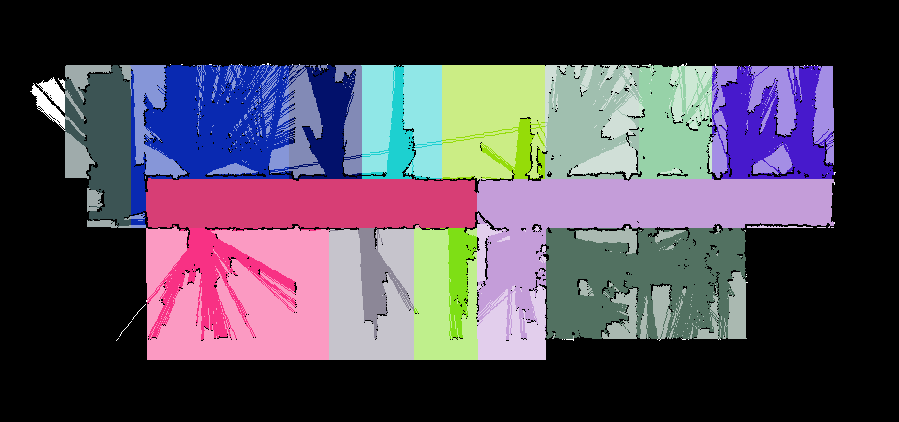}}
 \caption{Representative lines (a), Voronoi Graph (b), and resulting floor plan $\mathcal{F}$ (c) in a case where structure is not fully observed in the map. \label{fig:expl}} 
\end{figure}

In cluttered environments, it can happen that neither side of a wall that divides two different rooms $r$ and $r'$ is directly observed, due to partial mapping or occlusion. In these cases, we cannot rely directly on walls to separate those two rooms and identify their structure. 

Instead, we consider the building topology by computing a Voronoi topological graph $G=(N,T)$ of the environment, as shown in Fig. \ref{fig:VORONOI}, using the method of \cite{IROS18}, where $N$ and $T$ are the set of nodes and edges, respectively. In this way, we can check, for each room $r$, that all nodes $N_r$ belonging to $r$ are connected to each other (i.e., the sub-graph $G_r$ containing all nodes $N_r$ is a connected graph). If this condition does not hold, we split accordingly the room $r$ into two rooms $r'$ and $r''$ according to the separated components of the graph $G_r$.
To estimate the shape of $r'$ and $r''$  we rely on the symmetry of the building. If a representative line $l$ that divides $G_{r'}$ from $G_{r''}$ exists, we use it to separate those rooms in the floor plan $\mathcal{F}$. If such a line does not exist, we divide the two rooms by using a line with the same direction of the dominant direction $\Psi$ that can separate nodes in $G_{r'}$ from $G_{r''}$. An example is shown in Fig. \ref{fig:expl}, where Fig. \ref{fig:explB} shows the representative lines, Fig. \ref{fig:VORONOI} shows the Voronoi segmentation, and Fig. \ref{fig:explA} shows the final result with the segmented rooms and the floor plan $\mathcal{F}$ where the added lines are shown.

\section{Experimental Evaluation}\label{sec:EXP}

\begin{figure*}
 \centering
    	\subfloat[\ours{} IoU$=95.07$ \label{fig:ARTA}]{ \includegraphics[width=0.24\linewidth]{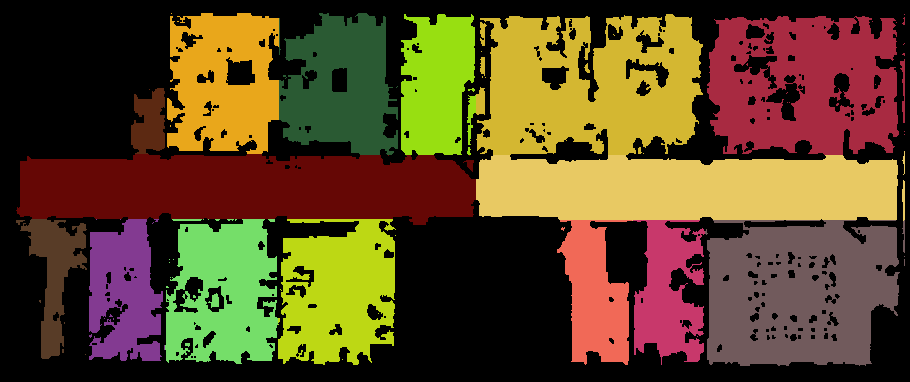}}
	\subfloat[\voronoi{} IoU$=27.34$ \label{fig:ARTB}]{ \includegraphics[width=0.24\linewidth]{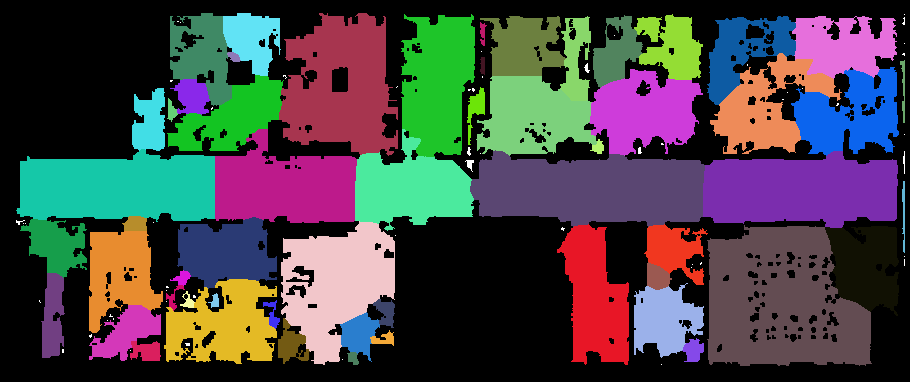}}
    	\subfloat[\morph{} IoU$=67.32$ \label{fig:ARTC}]{ \includegraphics[width=0.24\linewidth]{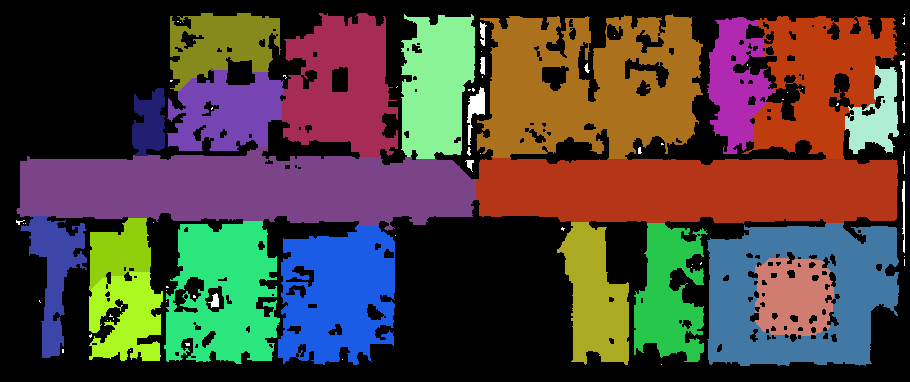}}	
    	\subfloat[\dist{} IoU$=63.39$ \label{fig:ARTD}]{ \includegraphics[width=0.24\linewidth]{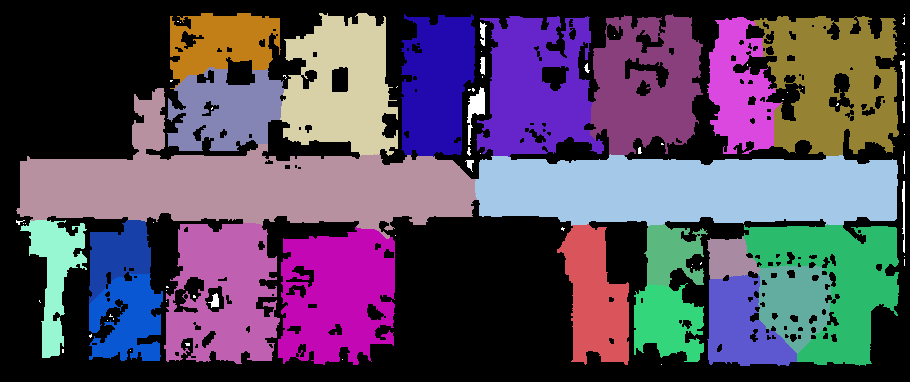}}\\
 \caption{Room segmentation of our method and of methods from \cite{bormann2016room} on the map of Fig. \ref{fig:EXE}.\label{fig:ART}} 
\end{figure*}

In this section we evaluate the capabilities of \ours{} to identifying semantically meaningful structure of cluttered 2D environments. 

To do so, we focus on the task of \emph{room segmentation}. 
Evaluation is performed both visually and quantitatively, comparing the segmented map $\check{M}$ with the actual ground-truth segmentation of the rooms $M_{Gt}$ (obtained with manual labelling and using the environment floor plan, when available, as a reference) of the same map $M$.
Given a room $\hat{r}$ obtained from $\check{M}$ and that of its ground-truth counterpart $r$ obtained from  $M_{Gt}$, we compute their \emph{Intersection over Union} (IoU) as $\text{IoU}(\hat{r},r) = \hat{r}\cap r / \hat{r} \cup r$.

Intuitively, we consider the area $r \setminus \hat{r}$ as a false negative, the area $\hat{r} \setminus r$ as a false positive, and the area of $\hat{r} \cap r$ as a true positive.
Then, given the segmented room $\hat{r}$ of a room in $\check{M}$, we look for the corresponding room $r$ in $M_{Gt}$ that maximizes the overlap with $\hat{r}$ using the same approach of~\cite{bormann2016room}. We scale IoU in range [0-100].

For comparison with \cite{bormann2016room}, we also report the metrics of \emph{precision} and \emph{recall}. Precision is defined as the maximum overlapping area of a segmented room with the corresponding ground truth room, divided by the area of the segmented room; recall is defined the maximum overlap of a ground truth room with the corresponding segmented room divided by the area of the ground truth room. Note that these two metrics have complementary bias \cite{mielle2018method}; an undersegmented map (fewer rooms than the actual ones) can have high recall and low precision; conversely, a oversegmented map (more rooms than the actual ones) can have high precision and low recall). We consider the IoU as the most relevant metric as it is not affected by this bias.

We compare our results (label \ours) against publicly available methods used in the room segmentation survey of  \cite{bormann2016room}: Voronoi-based (\voronoi ), morphological (\morph ), and distance-based segmentation (\dist ). 
As these methods, that are described in Section \ref{sec:ART}, are based on features extracted directly from the map $M$, their performance varies when the maps contain significant noise and clutter. Full results from  all considered maps and the code of the implemenentation of our method are available online\footnote{\url{https://github.com/goldleaf3i/declutter-reconstruct}}.
%\red{Precision and Recall are subject to score well over and undersegmentation. IoU is a more balanced metric.}

\subsection{Results on publicly available maps}\label{sec:pub}

\begin{table}[t]
\centering
\begin{adjustbox}{max width=\linewidth}
\begin{tabular}{lcccccccc}
\hline
            &  \ours  &  \morph &  \dist  &  \voronoi  \\
\hline
 precision  &   \textbf{89.02} (8.39)  &     84.98 (6.12)  &   88.34 (7.85)  &  85.86   (9.84)  \\
 recall     &    \textbf{93.93} (4.21)  &     82.3  (11.16) &   79.79 (12.15) &  71.92   (13.25)  \\
 IoU        &    \textbf{73.3}  (17.83) &     51.24 (12.25) &   54.65 (14.73) &  28.65   (10.55)  \\
\hline
\end{tabular}
\end{adjustbox}
\caption{Average results over 10 cluttered maps.}
\label{tab:new_sample}
\end{table}

In this section, we present the results obtained in 10 fully-explored maps available from the datasets of \cite{radish,strands}, as the one of Fig. \ref{fig:EXE}. Those maps are processed by \ours{} regardless of the algorithm used for performing SLAM and of their size, and without changing any parameter. As those maps are obtained in real environments, like offices, they naturally present a certain degree of clutter and artefacts.

\begin{figure}
 \centering
    	\subfloat[\ours{}\\ IoU$=80.44$\label{fig:101A}]{ \includegraphics[width=0.24\linewidth]{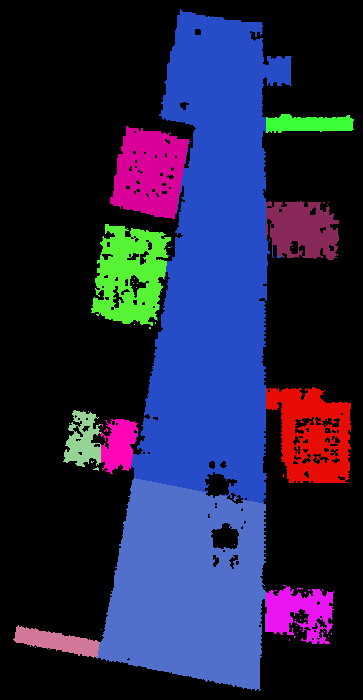}}
	\subfloat[\voronoi{}\\ IoU$=44.22$ \label{fig:101B}]{ \includegraphics[width=0.24\linewidth]{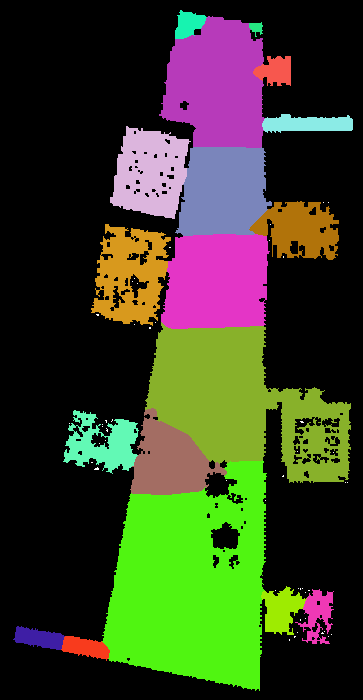}}
    	\subfloat[\morph{}\\ IoU$=55.65$ \label{fig:101C}]{ \includegraphics[width=0.24\linewidth]{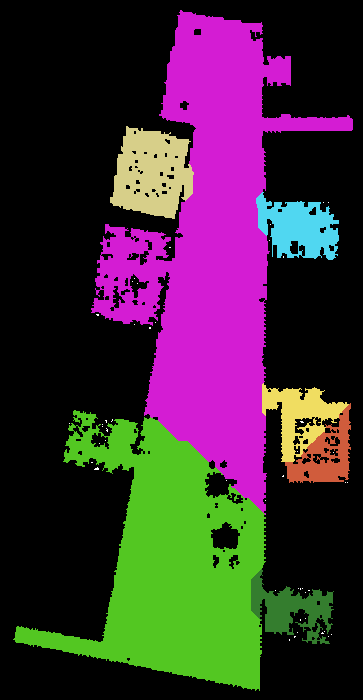}}	
    	\subfloat[\dist{} \\IoU$=52.39$ \label{fig:101D}]{ \includegraphics[width=0.24\linewidth]{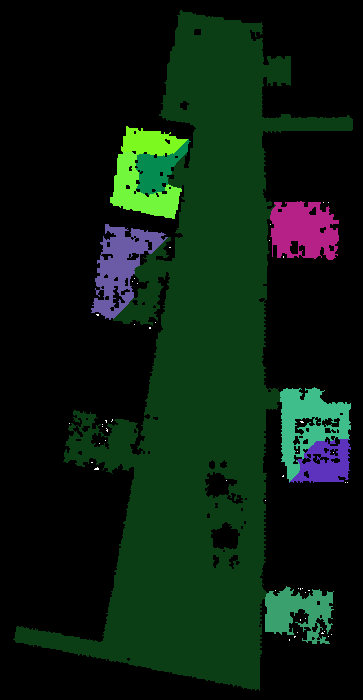}}\\
 \caption{Room segmentation of our method and of methods from \cite{bormann2016room} on a non-Manhattan map. \label{fig:101}} 
\end{figure}

Table \ref{tab:new_sample} presents the results. Our method clearly outperforms the other ones from \cite{bormann2016room}. The increase in performance is particularly clear when we consider the IoU, which is the metric that better describes how our method can produce a robust and stable segmentation.
Fig. \ref{fig:ART} compares the results obtained by our method with those obtained by other methods in segmenting the map of Fig. \ref{fig:EXE}, while Fig \ref{fig:101} shows results in a map with a non-Manhattan floor plan. In both examples, not only our method performs better than the others in terms of IoU, but also the errors of our method in segmenting the map are less critical, as the segmentation is coherent with the structure of the environment. At the same time, the methods of \cite{bormann2016room} often segment the environment in a way that is not semantically coherent with its structure, as for example in Figs. \ref{fig:ARTB} and \ref{fig:101C}.

\begin{figure}
 \centering
    	\subfloat[\ours{} IoU$=38.15$ \label{fig:withA}]{ \includegraphics[width=0.49\linewidth]{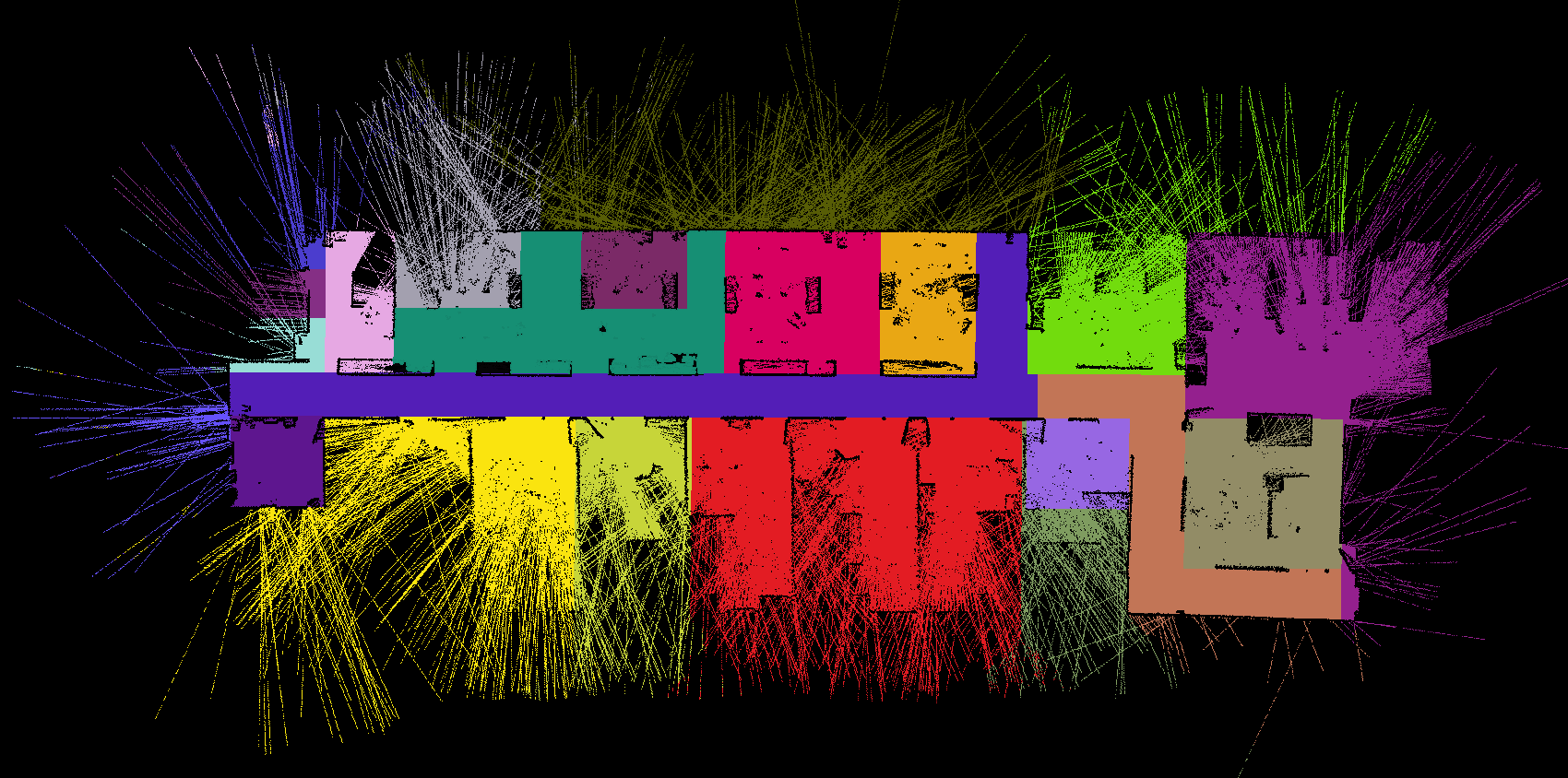}}
	\subfloat[\voronoi{} IoU$=10.23$ \label{fig:withB}]{ \includegraphics[width=0.49\linewidth]{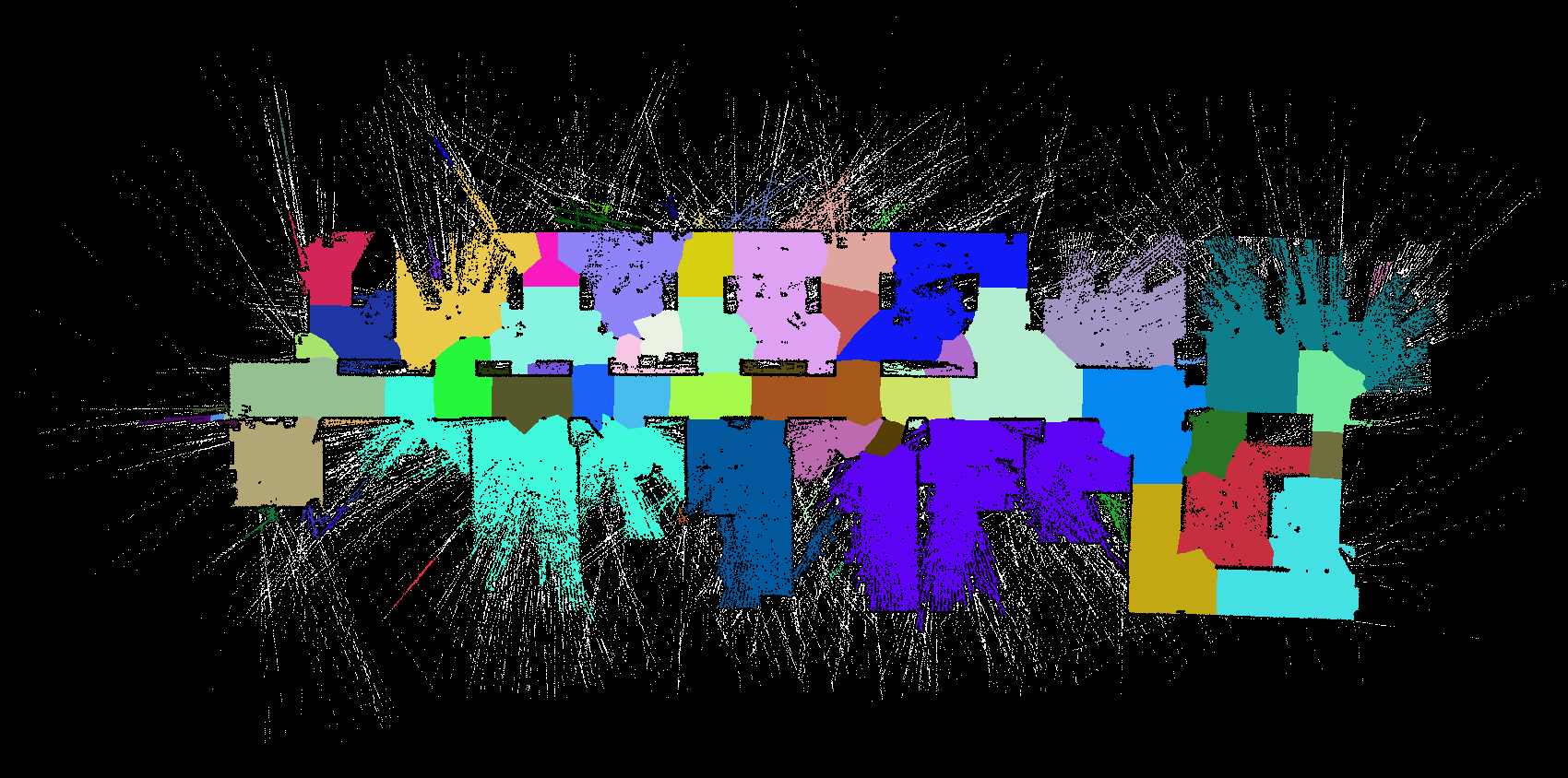}}\\
    	\subfloat[\morph{} IoU$=33.03$ \label{fig:withC}]{ \includegraphics[width=0.49\linewidth]{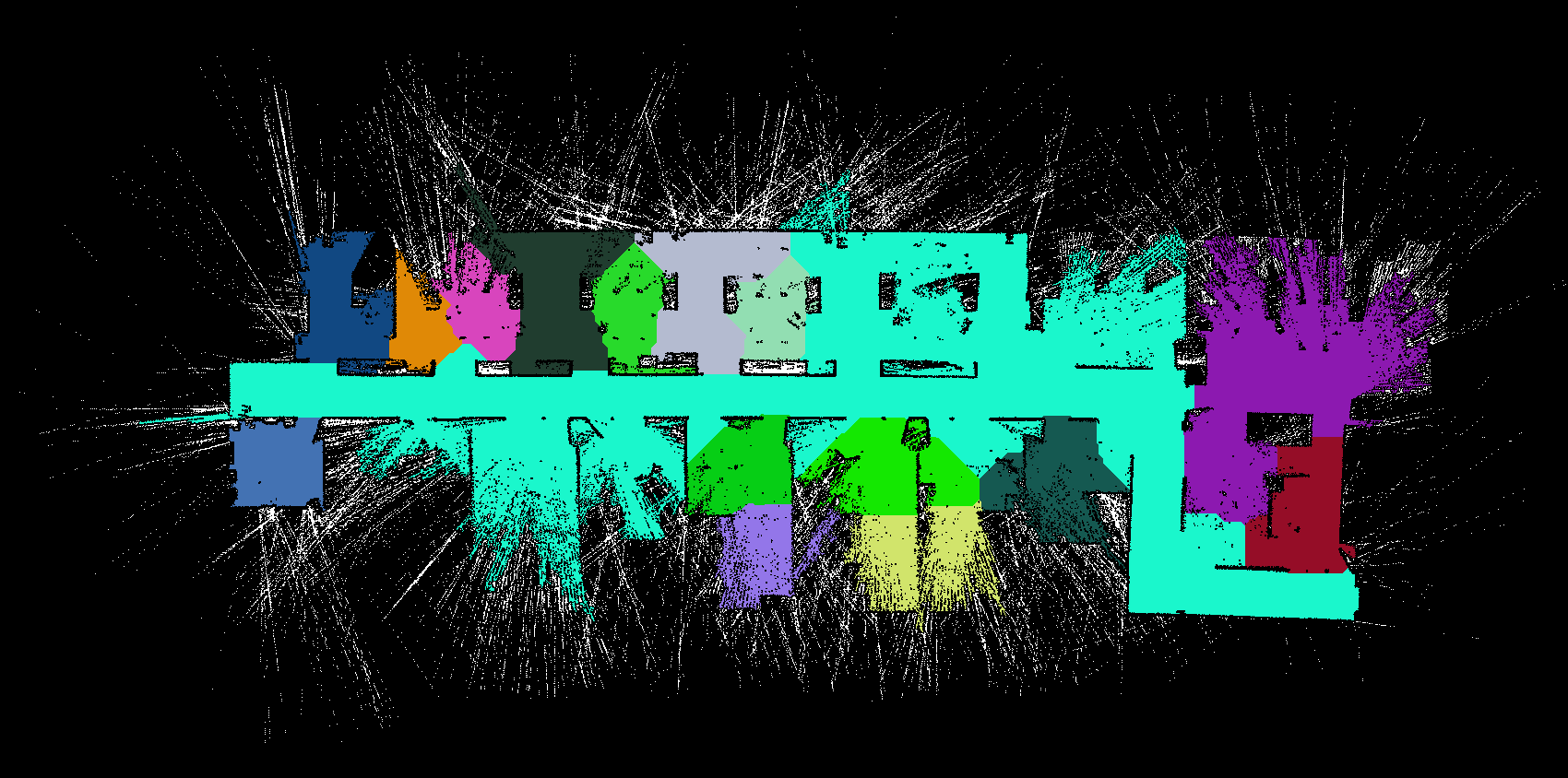}}	
    	\subfloat[\dist{} IoU$=39.31$ \label{fig:withD}]{ \includegraphics[width=0.49\linewidth]{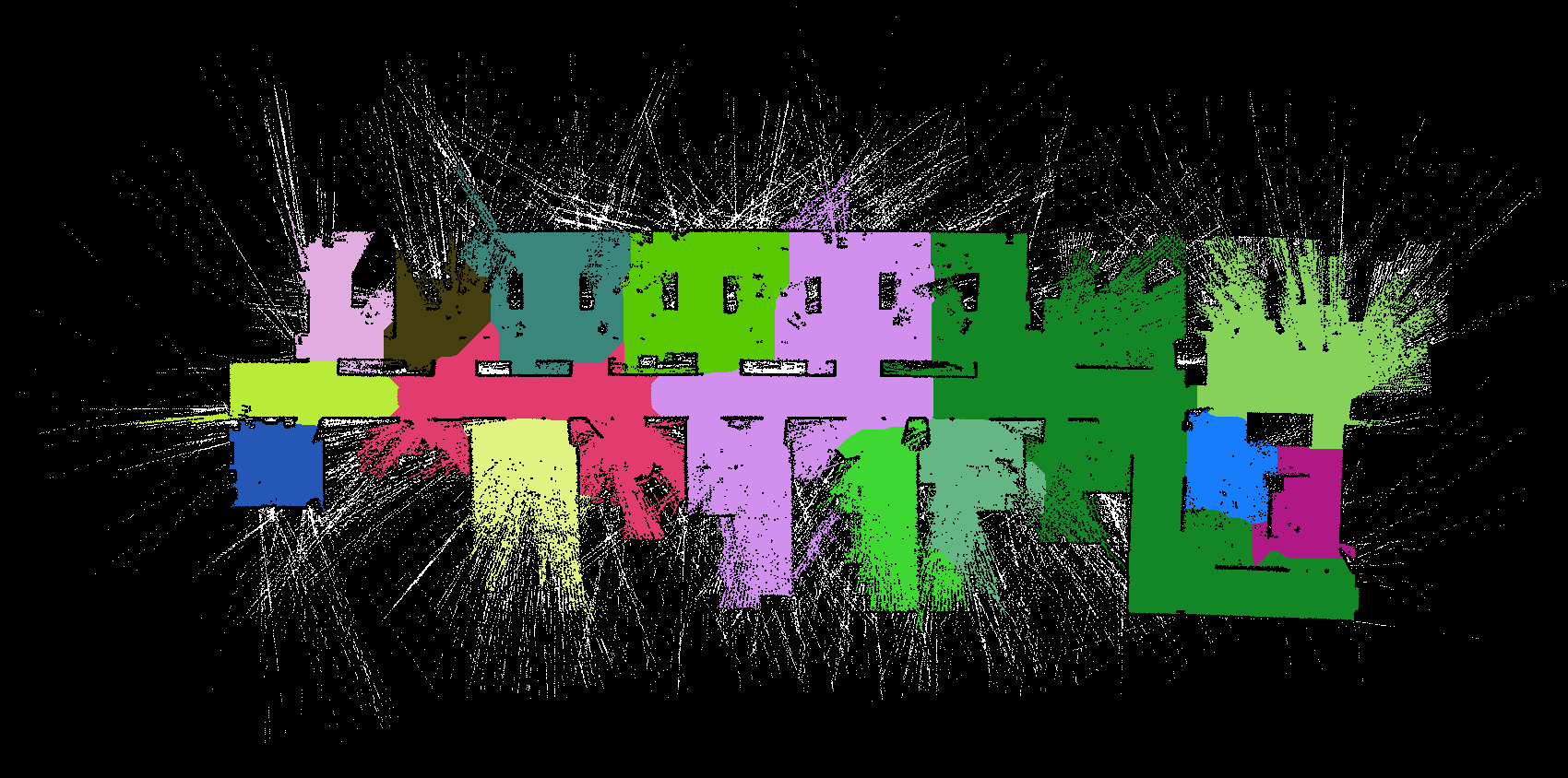}}\\
 \caption{Room segmentation of our method and of methods from \cite{bormann2016room} on a map with several perception artifacts.\label{fig:with}} 
\end{figure}

Fig. \ref{fig:with} shows results in a map where several artifacts due to perception issues (e.g., glass walls) are clearly visible. Nevertheless, our method, differently from those of \cite{bormann2016room}, compensates for those map inaccuracies and provides a meaningful segmentation of the environment, identifying its structure. 

To test the robustness and performance stability of \ours , we report the results on the dataset used as a benchmark in the survey on room segmentation methods of \cite{bormann2016room}. The dataset is composed of $20$ clean maps and $20$ maps with artificial additional noise of large-scale structured environments\footnote{\url{http://wiki.ros.org/ipa_room_segmentation}}. Those maps are significantly different (and simpler) from real-world maps, as they do not present clutter nor noise typically present in real-world maps, but are empty environments (\texttt{no furniture}) or with geometric noise added (\texttt{furnished}).

On these maps, \ours{} obtains results that are similar to those obtained in cluttered maps (and that are better than those of \cite{bormann2016room} and of our previous work of \cite{ICRA2019}), with a 93.54 (5.46) precision and 91.03 (3.46) recall on \texttt{furnished} maps, and 93.26 (6.4) precision and 97.58 (2.2) recall on \texttt{no furniture}, without changes in parameters. Full results are reported in the repository.

\subsection{Results on partial maps}\label{sec:part}

\begin{figure}
 \centering
     	\subfloat[\ours{} IoU$=80.98$\label{fig:explC0}]{ \includegraphics[width=0.495\linewidth]{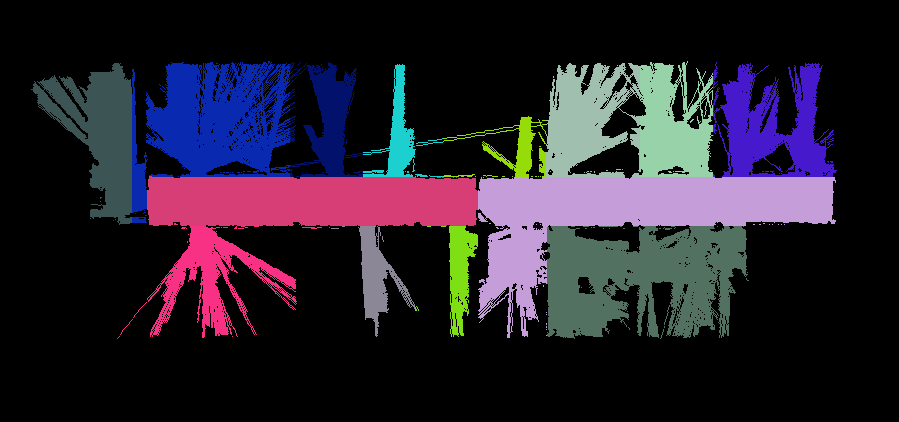}}
    	\subfloat[\voronoi{} IoU$=33.1$ \label{fig:explCA}]{ \includegraphics[width=0.495\linewidth]{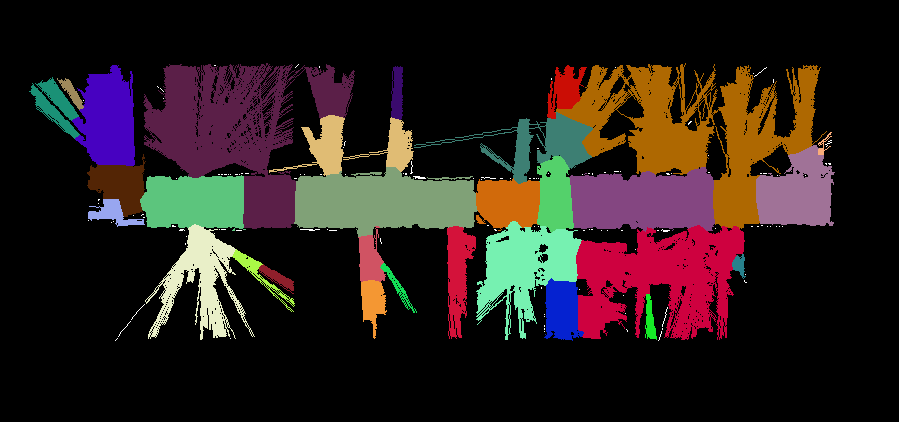}}\\
	\subfloat[\morph{} IoU$=66.47$ \label{fig:explCB}]{ \includegraphics[width=0.495\linewidth]{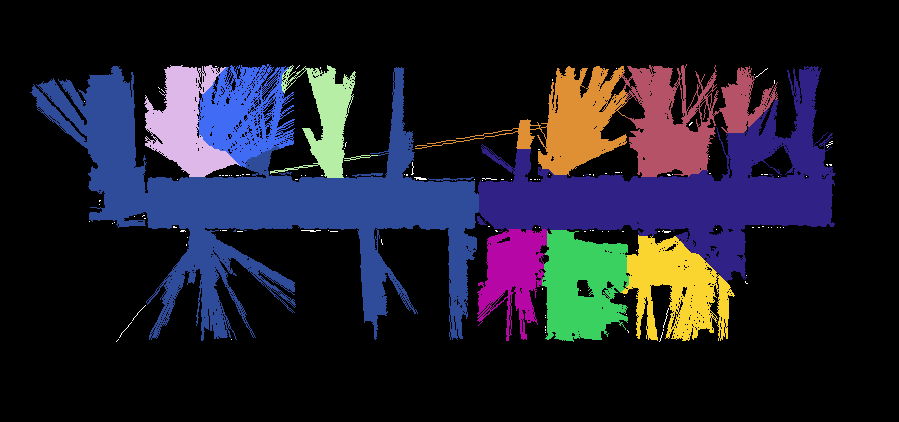}}
    	\subfloat[\dist{} IoU$=74.62$ \label{fig:explCC}]{ \includegraphics[width=0.495\linewidth]{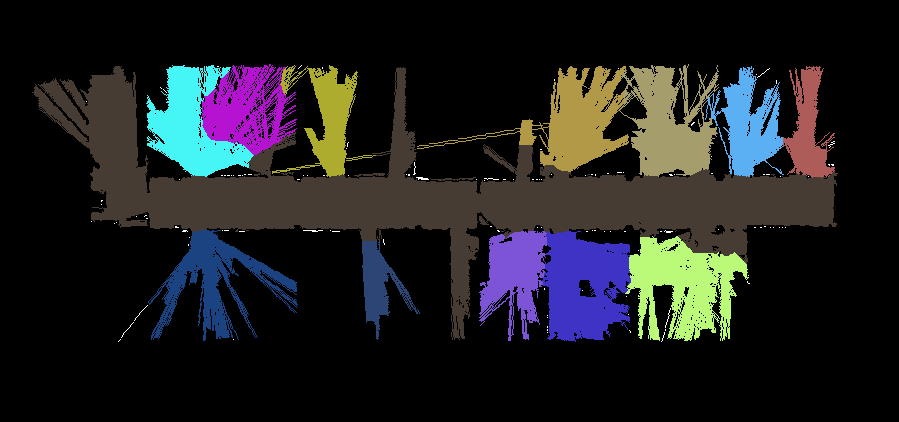}}	
 \caption{Room segmentation of our method and of methods from \cite{bormann2016room} on a partial map obtained during exploration.\label{fig:explC}} 
\end{figure}

In this section, we evaluate the results obtained in several maps coming from the incremental exploration of a building. In this way we test the robustness of \ours{} to extract the structural features from different types of maps. For example, maps obtained in the early stages of exploration have few rooms that are only partially mapped by the robot.  
Data are taken from \cite{radish} using two robot runs obtained in the  Freibug Building 79 (\texttt{FR79}) and in the University of Bremen Cartesium building (\texttt{Cartesium}). We relied on GMapping \cite{gmapping2007tro} as SLAM method. We used the last map to compute the coverage percentage of the map from the start of exploration (coverage = 0) to the full exploration of the environment (coverage = 1).

Fig. \ref{fig:explC} shows the results of \ours{} compared with the methods of \cite{bormann2016room} in a partial map obtained in \texttt{FR79}. \ours{} can segment the environment in a meaningful way also when the structure of the building can be identified by only a few walls as partially perceived by the robot. Conversely, the results of the methods from \cite{bormann2016room} show several artifacts. Fig. \ref{fig:expl} shows the features we extract, namely the floor plan $\mathcal{F}$ and representative lines $l$, obtained in the same environment of Fig \ref{fig:explC}. Note how  $\mathcal{F}$ is a reliable representation of the actual shape of the environment, which could be used by the robot to have a better understanding of the shape of rooms which are only partially perceived by the robot.

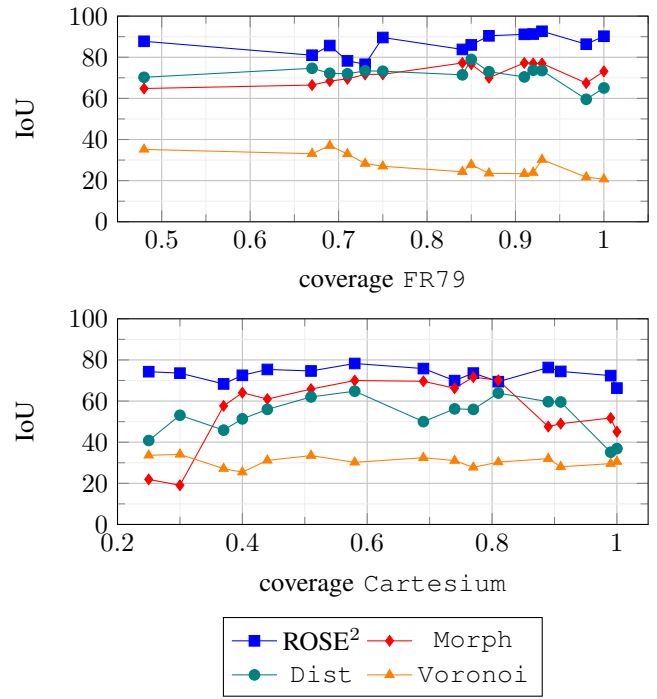
\begin{figure}
 \centering
\begin{tikzpicture}
\begin{axis}[
    xmin = 0.45, xmax = 1.05,
    ymin = 0, ymax = 100,,
  ylabel=IoU,
  xlabel= coverage \texttt{FR79},    
  grid = both,
  minor tick num = 1,
  major grid style = {lightgray},
  minor grid style = {lightgray!25},
  width = \columnwidth,
  height = 0.5\columnwidth]
\addplot[blue, mark = square*] table [x=$COV$, y=$geom$]{FR79.dat};
%\addlegendentry{$geom$ series}
\addplot[red, mark = diamond*] table [x=$COV$, y=$mor$]{FR79.dat};
%\addlegendentry{$mor$ series}
\addplot[teal, mark =*] table [x=$COV$, y=$dist$]{FR79.dat};
%\addlegendentry{$dist$ series}
\addplot[orange, mark = triangle*] table [x=$COV$, y=$vor$]{FR79.dat};
%\addlegendentry{$vor$ series}
\end{axis}
%\node[above] at (current bounding box.north) {\texttt{fr79}};
\end{tikzpicture}
\begin{tikzpicture}
\begin{axis}[
xmin = 0.2, xmax = 1.05,
    ymin = 0, ymax = 100,
  xlabel=coverage \texttt{Cartesium},
  ylabel=IoU,    
  grid = both,
  minor tick num = 1,
  major grid style = {lightgray},
  minor grid style = {lightgray!25},
  width = \columnwidth,
  height = 0.5\columnwidth,
   legend columns=2, 
  legend style={at={(0.5,-0.45)},anchor=north}]
\addplot[blue, mark = square*] table [x=$COV$, y=$geom$]{BREMEN.dat};
\addlegendentry{ \ours }
\addplot[red, mark = diamond*] table [x=$COV$, y=$mor$]{BREMEN.dat};
\addlegendentry{ \morph }
\addplot[teal, mark =*] table [x=$COV$, y=$dist$]{BREMEN.dat};
\addlegendentry{ \dist{}  }
\addplot[orange, mark = triangle*] table [x=$COV$, y=$vor$]{BREMEN.dat};
\addlegendentry{ \voronoi }
\end{axis}
{\texttt{Cartesium}};
\end{tikzpicture}
 \caption{IoU vs.\ percentage of coverage for the \texttt{FR79} (top) \texttt{Cartesium} (bottom).\label{fig:INCREMENTAL}}
 \vspace{-0.5cm}
\end{figure}

Fig. \ref{fig:INCREMENTAL} shows the IoU at different coverage levels for the two environments \texttt{FR79} and \texttt{Cartesium}. The performance of our methods are consistent (IoU around 80) at all degrees of map completion. At the same time, qualitatively speaking, the segmentation performed by other methods varies greatly as the map of the environment changes, while also achieving a lower quantitative performance.

\subsection{Discussion }

The results from Fig. \ref{fig:INCREMENTAL} show how our method is robust in different map conditions; this is due to the fact that the steps described in Section \ref{sec:OUR} allow the extraction of semantically meaningful structural features even from  partial maps or from particularly cluttered ones.

\begin{figure}
   \centering
  \includegraphics[width=0.495\linewidth]{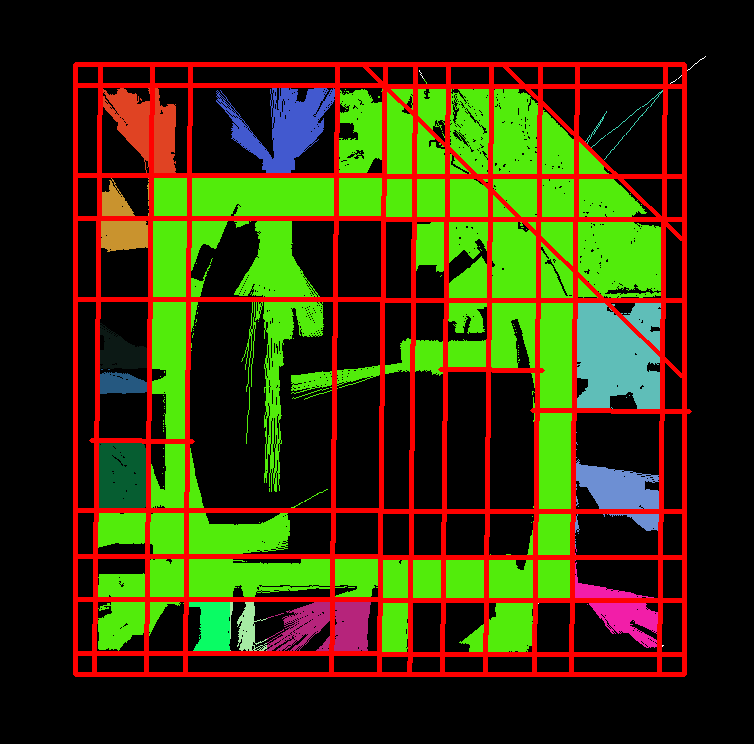}
  \caption{ 
Features extracted (representative lines, room segmentation) from a partial map of \texttt{INTEL} (IoU$=75.98$).}
  \label{fig:INTEL}
\end{figure}

\begin{figure}
   \centering
   \subfloat[IoU$=90.42$\label{fig:predA}]{
  \includegraphics[width=0.45\linewidth]{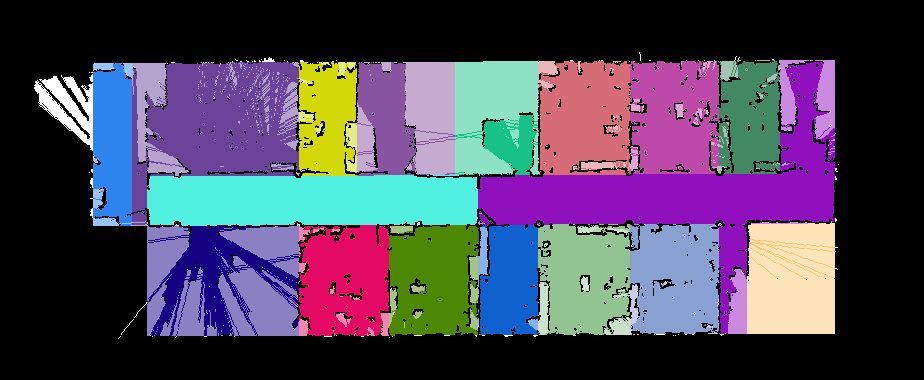}}
  \subfloat[IoU$=74.63$ \label{fig:predB}]{
  \includegraphics[width=0.52\linewidth]{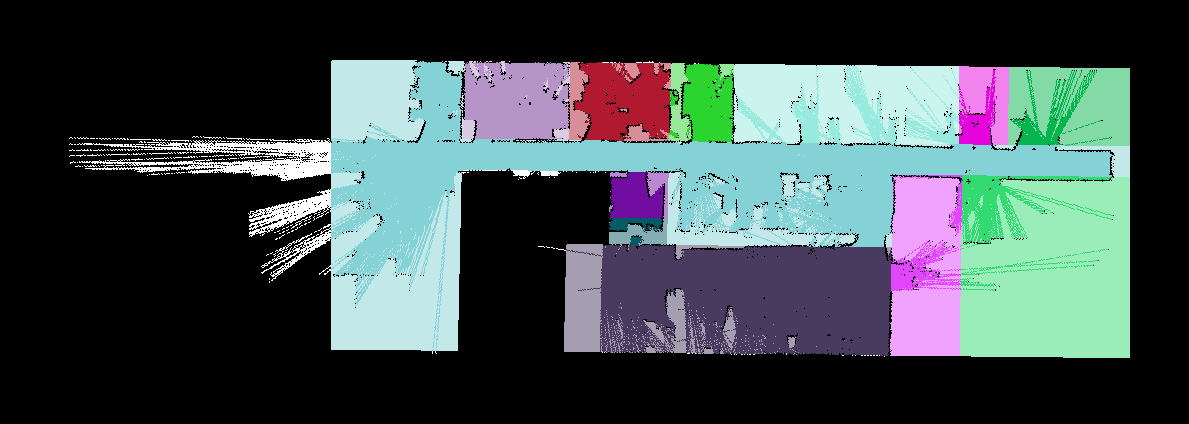}}
  \caption{ 
Floor plan $\mathcal{F}$ and room segmentation of two partial maps from \texttt{FR79} (a) and \texttt{Cartesium} (b).}
  \label{fig:PRED}
  \vspace{-1em}
\end{figure}

Another example of this is in Fig. \ref{fig:INTEL}, where we show the segmentation and the representative lines retrieved from a partial map obtained from the Intel Lab (\texttt{INTEL}) \cite{radish}. Also in this environment, which is complex due to its large size and its non-Manhattan layout, our method gives a semantically meaningful reconstruction of the environment, while providing the directions of the main walls.

The retrieval of the floor plan $\mathcal{F}$ and the identification of walls and representative lines can provide meaningful insights on the actual shape of the environment also in cases the robot has not a full knowledge of the entire map. An example is shown in Fig. \ref{fig:PRED} with the inpainted room shape from the floor plan $\mathcal{F}$ for two partial maps obtained from \texttt{FR79} and \texttt{Cartesium}. Despite that there are several parts of the environment not observed yet by the robot, our method can give an accurate representation of its structure. This knowledge could be used to complement the map, by estimating the actual shape of the rooms,  but also to give extra knowledge and awareness of the environment to the robot \cite{AAMAS21}.

\section{Conclusions}\label{sec:CON}
In this work, we have presented a method to extract the structure of an environment from its (cluttered) 2D occupancy grid map, while performing a segmentation of the map into a set of rooms.
Our method starts to identify the main directions of lines as observed in the occupancy map, and cleans the map from clutter and noise. After that, it estimates the presence of the walls in the environment and of their directions with a set of representative lines; these lines are combined to obtain a geometrical floor-plan-like representation that is used to segment the map in rooms. 
Results show that our method can be successfully applied to occupancy grid maps, even in the case of severe noise and clutter and when the map represents only a part of the environment, as during exploration. 
Future work involves the use of structural information to improve long-term mapping of changing environments.
%When compared with state-of-the-art methods used for room segmentation, we can obtain better and more stable performances in challenging settings. At the same time, the structure of the environment present a coherence with the one of the environment.

%%%%%%%%%%%%%%%%%%%%%%%%%%%%%%%%%%%%%%%%%%%%%%%%%%%%%%%%%%%%%%%%%%%%%%%%%%%%%%%%


%%%%%%%%%%%%%%%%%%%%%%%%%%%%%%%%%%%%%%%%%%%%%%%%%%%%%%%%%%%%%%%%%%%%%%%%%%%%%%%%

%%%%%%%%%%%%%%%%%%%%%%%%%%%%%%%%%%%%%%%%%%%%%%%%%%%%%%%%%%%%%%%%%%%%%%%%%%%%%%%%

%%%%%%%%%%%%%%%%%%%%%%%%%%%%%%%%%%%%%%%%%%%%%%%%%%%%%%%%%%%%%%%%%%%%%%%%%%%%%%%%
%\section*{APPENDIX}


%\section*{ACKNOWLEDGMENT}

%The preferred spelling of the word ÒacknowledgmentÓ in America is without an ÒeÓ after the ÒgÓ. Avoid the stilted expression, ÒOne of us (R. B. G.) thanks . . .Ó  Instead, try ÒR. B. G. thanksÓ. Put sponsor acknowledgments in the unnumbered footnote on the first page.

%%%%%%%%%%%%%%%%%%%%%%%%%%%%%%%%%%%%%%%%%%%%%%%%%%%%%%%%%%%%%%%%%%%%%%%%%%%%%%%%
\bibliographystyle{IEEEtran}
\bibliography{citations}

\end{document}